\documentclass[journal]{IEEEtran}
\usepackage{amsmath,amssymb}

\usepackage[caption=false]{subfig}
\usepackage{multirow}
\usepackage{booktabs}
\usepackage{overpic}
\usepackage{color}
\usepackage[outdir=./]{epstopdf}

\makeatletter
\newif\if@restonecol
\makeatother

\usepackage[linesnumbered,ruled,vlined]{algorithm2e}
\usepackage{algpseudocode}
\SetKw{Continue}{continue}
\SetKw{Return}{return}

\begin{document}
\title{A CRF-based Framework for Tracklet Inactivation in Online Multi-Object Tracking}
\author{Tianze Gao, ~Huihui Pan, ~Zidong Wang, ~\IEEEmembership{Fellow,~IEEE}, ~and ~Huijun Gao, ~\IEEEmembership{Fellow,~IEEE}
\thanks{Manuscript received October 21, 2020; revised January 21, 2021; accepted February 22, 2021. This work was supported in part by the National Natural Science Foundation of China under Grant U1964201, Grant 61803120, and Grant 61790562. The associate editor coordinating the review of this manuscript and approving it for publication was Dr. Marco Carli. (\textit{Corresponding author: Huijun Gao.})}
\thanks{Tianze Gao and Huijun Gao are with the Research Institute of Intelligent Control and Systems, Harbin Institute of Technology, Harbin, 150001, China (e-mail: gao2990026796@gmail.com; hjgao@hit.edu.cn).}
\thanks{Zidong Wang is with the Department of Computer Science, Brunel University London, Uxbridge, Middlesex, UB8 3PH, United Kingdom (e-mail: zidong.wang@brunel.ac.uk).}
\thanks{Huihui Pan is with the Research Institute of Intelligent Control and Systems, Harbin Institute of Technology, Harbin, 150001, China and also with the Ningbo Institute of Intelligent Equipment Technology Company, Ltd, Ningbo 315200, China (e-mail: huihuipan@hit.edu.cn).}
\thanks{Color versions of one or more figures in this article are available online at https://10.1109/TMM.2021.3062489.}
\thanks{Digital Object Identifier 10.1109/TMM.2021.3062489}
}
\maketitle

\begin{abstract}
Online multi-object tracking (MOT) is an active research topic in the domain of computer vision. Although many previously proposed algorithms have exhibited decent results, the issue of tracklet inactivation has not been sufficiently studied. Simple strategies such as using a fixed threshold on classification scores are adopted, yielding undesirable tracking mistakes and limiting the overall performance. In this paper, a conditional random field (CRF) based framework is put forward to tackle the tracklet inactivation issue in online MOT problems. A discrete CRF which exploits the intra-frame relationship between tracking hypotheses is developed to improve the robustness of tracklet inactivation. Separate sets of feature functions are designed for the unary and binary terms in the CRF, which take into account various tracking challenges in practical scenarios. To handle the problem of varying CRF nodes in the MOT context, two strategies named as hypothesis filtering and dummy nodes are employed. In the proposed framework, the inference stage is conducted by using the loopy belief propagation algorithm, and the CRF parameters are determined by utilizing the maximum likelihood estimation method followed by slight manual adjustment. Experimental results show that the tracker combined with the CRF-based framework outperforms the baseline on the MOT16 and MOT17 benchmarks. The extensibility of the proposed framework is further validated by an extensive experiment.
\end{abstract}


\begin{figure*}[htbp]
	{\centering
	\includegraphics[width=0.9\textwidth]{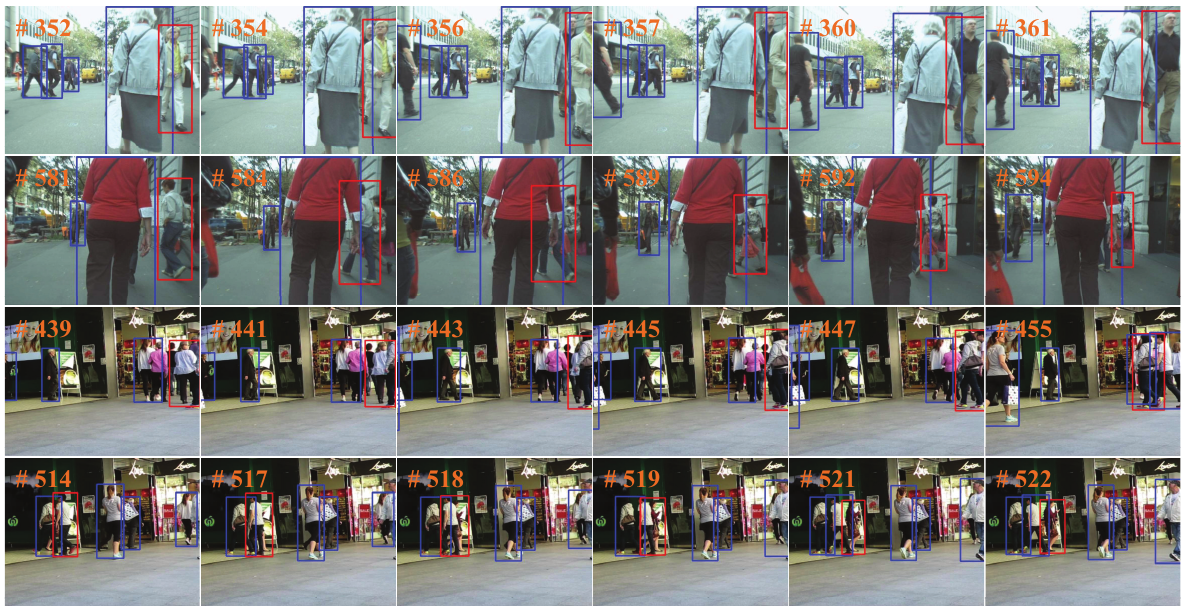}

	\caption{Examples of the improper tracklet inactivation in Tracktor++. The tracking hypotheses in red boxes drift to nearby objects as a consequence of not being correctly inactivated. This is a common problem happening in trackers without a dedicated algorithm for tracklet inactivation.}
	\label{fig1}
}
\end{figure*}

\section{Introduction}
\label{sec:introduction}
\IEEEPARstart{M}{ulti-object} tracking (MOT) has become a popular research topic attracting an ever-increasing interest from the computer vision community due to its wide applying prospect. In general, MOT algorithms are categorized into online algorithms and offline ones according to whether the information from future frames is incorporated in data association. Compared with the offline counterparts, online algorithms are more suitable for real time applications including autonomous driving and mobile robotics. In this paper, we focus on online MOT algorithms.

The current mainstream online MOT paradigm is tracking-by-detection, which presents a two-step solution to the online MOT problem: 1) discover objects of interest through a detector; and 2) form trajectories by using the data association method. It should be mentioned that the tracking-by-detection paradigm is heavily dependent on the detection quality, and an additional feature extractor is sometimes required to compute the appearance affinity. Another paradigm for online MOT is tracking-by-SOT, wherein SOT stands for the single object trackers. Some correlation filter based MOT tracking methods~\cite{Chu2019a,Zhu2018b} fall into this scope.

Bearing the basic ideas of tracking-by-detection, some researchers have further proposed to simultaneously detect and track objects within a unified framework~\cite{Bergmann2019,zhou2020tracking}. The regression ability of convolutional neural networks (CNN) has been leveraged to estimate the offsets of detections across adjacent video frames. Methods of this type have proven to be effective by state-of-the-art online trackers Tracktor++~\cite{Bergmann2019} and CenterTrack~\cite{zhou2020tracking}. However, there is still leeway for improvement regarding the strategy for tracklet inactivation. For online MOT problems, inactivating a tracklet means that a tracklet is temporarily removed from the current tracklet list while keeping its potential qualification to be reidentified. Most current online MOT algorithms would predefine a life span for each tracklet to determine the maximal number of continuous inactivation frames before it is thoroughly discarded.

Previous works have generally resorted to simple strategies on tracklet inactivation. For example, Tracktor++~\cite{Bergmann2019} inactivates tracklets by setting a fixed threshold on the classification scores of tracking hypotheses. DMAN~\cite{Zhu2018b} sets two inactivation thresholds on both confidence scores and overlap ratios. CenterTrack~\cite{zhou2020tracking} inactivates the tracklets that fail to find company in the greedy matching stage. These strategies are limited in exploiting the inter-object relationships and thus lead to certain tracking failures, especially those related to tracklet identification (ID).

In this paper, a CRF-based framework is presented for handling tracklet inactivation in online MOT methods. A discrete CRF model is developed, with dedicated feature functions designed to cope with various tracking challenges. As an example of the applying targets, the Tracktor++ is taken as the baseline to which the proposed framework is applied. It will be shown that the drawback of tracklet inactivation in Tracktor++ (as is exhibited in Fig.~\ref{fig1}) can be considerably addressed by integrating the proposed framework. Moreover, since the CRF-based framework is decoupled from the main tracking pipeline, it can be naturally transferred to other trackers which are short of dedicated mechanism for tracklet inactivation. For the purpose of showing this kind of extensibility, the proposed framework is further adapted to CenterTrack as an extensive experiment.

There are two emerging concerns when using a CRF in the MOT context: \romannumeral1) the number of CRF nodes is varying according to the detected objects in a certain frame, whereas the traditional formulation of the CRF is built upon a fixed set of nodes; and \romannumeral2) the overfitting problem may occur during the training process under the condition that the scale of the training dataset is relatively small with respect to the number of CRF parameters. In this context, two strategies are adopted to address the aforementioned concerns. On the one hand, the hypothesis filtering and dummy nodes are applied to fix the number of CRF nodes. On the other hand, the graph factors are divided into the unary and binary groups. Then, the parameter sharing strategy is utilized within each of the two groups, which effectively reduces the risk of overfitting.

In the proposed framework, the tracking objects in the scene are treated equally (i.e., no object is considered to be more special than others), and thus, the CRF is constructed in a fully connected way. The loopy belief propagation algorithm proposed in \cite{Kyburg1991} is applied for the inference phase. For the training phase, the maximum likelihood estimation method is adopted and the stochastic gradient descent (SGD) algorithm is utilized to update the parameters.

The main contributions of this paper can be summarized as follows:
\begin{itemize}
	\item A CRF-based framework is developed for handling tracklet inactivation, with a state-of-the-art tracker as the baseline to give an example of its applying prospect.
	\item Dedicated feature functions for unary and binary terms are designed to cope with multiple concerns encountered in practical tracking scenarios.
	\item Experiments conducted on the MOT16 and MOT17 datasets demonstrate the superiority of the tracker refined by our framework over the baseline tracker.
	\item The extensibility of the proposed framework is validated by an extensive experiment.
\end{itemize}

The rest of the paper is organized as follows. The background of the MOT problem is introduced in Section \ref{sec:related_work}. In Section \ref{sec:method}, the details of the proposed framework are presented. The experimental settings are discussed in Section \ref{sec:implementation_details}. The details of performance indicators, experimental results, ablation study and an extensive experiment are provided in Section \ref{sec:experiments}. An overall conclusion is given in Section \ref{sec:conclusion}.

\section{Related work}
\label{sec:related_work}
During the past few decades, MOT has become a research focus in computer vision. In this section, we briefly review the MOT methods that leverage the power of machine learning and those with the aid of CRF.

\subsection{MOT with Machine Learning}
\label{subsec:mot_with_deep_learning}
Machine learning approaches have proven to be effective for dealing with the MOT problems through years of practice. The most popular paradigm in this direction is tracking-by-detection \cite{Ciaparrone2019,Mekonnen2019}. The interested objects are firstly detected by using a deep convolutional neural network. The affinity scores based on appearance and motion information of the objects are then computed. Finally, the objects are correlated through data association. One of the first algorithms that follows this paradigm is the Simple Online and Realtime Tracking (SORT) algorithm \cite{Bewley2016a} which uses the Kuhn–Munkres algorithm \cite{Kuhn1955a} for data association and a Kalman filter \cite{Kalman1960b} for observation correction. DeepSORT has been proposed in \cite{Wojke2018b} to further improve SORT by learning a deep association metric in visual appearance space. Zhou et al. \cite{zhou2018deep} have enhanced the tracking accuracy by using a deep alignment network. In addition, a local-to-global strategy for robust data association has been introduced in \cite{dai2018instance}. In recent years, some dedicated tracking algorithms have been designed either to increase the recall and accuracy of the detector, or to enhance the performance of data association \cite{bao2020pose,Yu2016,Karunasekera2019,fu2019multi,Lu2017,Sheng2019a}.

Another line of works follow the paradigm of tracking-by-SOT, which assigns an individual single object tracker for each tracking target. The approaches following this paradigm are effective because single object trackers often pay more attention to the similarity between nearby image patches in consecutive frames than to the integrity of semantic information. Hence, some objects that the detector fails to discover can be recognized by using the single object trackers. This point of view holds particularly well in the trackers based on correlation filters. For example, the kernelized correlation filters (KCF) proposed in \cite{Henriques2015b} have been employed as single object trackers in \cite{Chu2019a}. In \cite{Zhu2018b}, the efficient convolution operators (ECOs) proposed in \cite{Danelljan2017b} have been utilized for dealing with the online MOT problem.

More recently, Bergmann et al.~\cite{Bergmann2019} have developed a simultaneous detection and tracking framework named as Tracktor++, which achieves state-of-the-art tracking performance. It leverages a Faster R-CNN to discover new tracking targets. The positions of confirmed tracking objects are regressed in later frames by using the regression head of the Faster R-CNN, which differs from the traditional tracking-by-detection approaches. Another state-of-the-art tracker, CenterTrack~\cite{zhou2020tracking} has adopted a similar notion. It takes the concatenation of two images as input and directly predicts the offsets of objects across adjacent frames. A greedy matching algorithm is then employed for subsequent data association.

Despite of the various merits possessed by previous methods, the issue of tracklet inactivation has not been sufficiently investigated. Trackers without reliable mechanism for tracklet inactivation have the drawback of being liable to make ID related mistakes. In this paper, we study and address this issue by developing a CRF-based framework.

\subsection{MOT with CRF}
\label{subsec:mot_with_CRF}
CRF models are frequently used in Natural Language Processing (NLP) and visual segmentation tasks. The advantage of CRF is its strong ability to model the complex interactions between individuals. Therefore, CRF has been utilized to solve MOT problems.

Most research works on CRF-based methods are geared towards offline MOT problems, i.e., information from future frames is available to be utilized. Yang et al. \cite{Yang2012a} have built a fully connected CRF model that associated short tracklets to form long ones. The CRF edges have been designed to focus on discriminating spatially close targets with similar appearances. Heili et al. \cite{heili2014} have exploited long-term connectivity between different detections. Multiple cues have been extracted to measure the similarity and dissimilarity when formulating the energy potentials. A recent work \cite{Xiang2019a} has followed the basic ideas of \cite{Yang2012a}, whereas it learns the pairwise potentials by using a bidirectional long short-term memory (LSTM) network, and approximated the inference phase of the CRF by using a recurrent neural network (RNN). The entire tracking pipeline is thus differentiable and can be trained end-to-end.

By contrast, the application of CRF in online MOT problems has not yet been fully studied. Zhou et al. \cite{Zhou2019} have proposed a method to solve the MOT problem in an online way. In \cite{Zhou2019}, the displacements between consecutive frames have been first estimated by using a dedicated neural network. Then, a deep continuous CRF with asymmetric pairwise terms has been utilized to refine the displacements. Note that our method is different from \cite{Zhou2019} in that we use a discrete CRF to judge the inactivation of tracklets instead of directly interacting with the positions of tracking hypotheses. Besides, since our framework is decoupled from the main tracking pipeline, it is featured with flexibility and extensibility in the sense of being applied to more advanced algorithms in the future.

\section{Method}
\label{sec:method}
In this section, we first make a brief description for the baseline tracker. Next, the formulation of the proposed CRF framework is presented. Then we show how the feature functions are designed by taking the interaction between different targets into account. Finally, the method for CRF inference and parameter estimation is introduced, including the two techniques employed to keep the number of CRF nodes fixed.
\subsection{Baseline Tracker}
For better understanding of the proposed framework, we deliver a short introduction of the baseline tracker, Tracktor++. As a tracking-by-detection method, Tracktor++ detects the tracking objects of interest through a Faster R-CNN detector. With the arrival of each new frame, the latest bounding box of each tracklet is treated as the region proposal and sent to the regression and classification heads. The new positions and scales of the tracklets are then provided by the output of the regression head. Once the classification score of a tracklet is lower than a predefined threshold, the tracklet is inactivated. The new detections that are covered by tracklets are filtered out by non-maximum suppression. An individual siamese network is also employed to reidentify the previously inactivated tracklets. Finally, the remaining detections are treated as the starting points of new tracklets.

\subsection{Problem Formulation}
\label{subsec:problem_formulation}
\begin{figure}[!t]
	\centering{\includegraphics[width=0.5\textwidth]{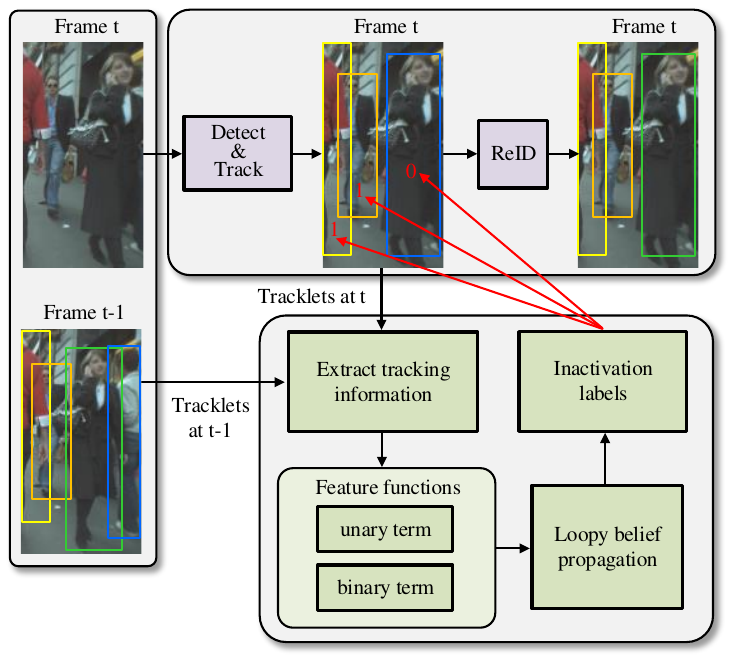}}
	\caption{The general pipeline of the proposed framework. We pose the judgement for tracklet inactivation as a binary labeling problem, where label $ 0 $ indicates that the tracklet should be inactivated and label $ 1 $ indicates the contrary. Given the tracklets at frame $t-1$ and frame $t$, the necessary information for calculating feature functions of the CRF model are first extracted. Then the inactivation labels are jointly determined by feature functions in the unary and binary terms through loopy belief propagation. In the presented example, the tracklet with the tracking hypothesis enclosed by the blue bounding box is correctly inactivated by the CRF framework, hence enabling its true identity to be recovered through the subsequent reidentification stage.}
	\label{fig2}
\end{figure}

The general pipeline of the proposed framework is illustrated in Fig.~\ref{fig2}, where the state-of-the-art method, Tracktor++, is used as the baseline tracker. Given a set of $ K $ regressing results $ x^t=\{x_1^t,x_2^t,\ldots,x_K^t\} $, our goal is to make a reasonable judgment $ y^t=\{y_1^t,y_2^t,\ldots,y_K^t\} $ labeling whether the tracklets should be inactivated. Here, we use $x_i^t=\{S_i^t,m_i^t,l_i^t\}$ as the ensemble tracking information of the tracklet $ i $ at frame $ t $, including the classification score $ S_i^t $, the motion information $ m_i^t $ and the size information $ l_i^t $. We will omit the frame index $ t $ for simplicity in the following formalism.$ x^t=\{x_1^t,x_2^t,\ldots,x_K^t\} $

We model the inactivation labeling problem with a CRF which is factorized as a factor graph \cite{Kschischang2001} $ G=(V,F,\mathcal{E}) $, where $ V=\{1,2,\ldots,K\} $ is the set of nodes corresponding to the tracklet indices. $ F $ and $ \mathcal{E} $ are the sets of factors and edges in the graph respectively. The joint conditional distribution can thus be expressed as
\begin{equation}
\label{eq1}
p(y|x;\theta)=\frac{1}{Z(x;\theta)}\prod_{f \in F}\text{e}^{-E_f(x,y;\theta)},
\end{equation}
with
\begin{equation}
\label{eq2}
Z(x;\theta)=\sum_{y\in Y} \prod_{f \in F}\text{e}^{-E_f(x,y;\theta)},
\end{equation}
where $ Y $ denotes the domain of $ y $. $ E_f $ is the energy function of the factor $ f $.

Note that the dataset exploited for training the CRF is established by using the improper inactivation instances produced by Tracktor++ as negative samples. Such instances simply produce a small scale dataset. Thereby, parameter sharing is applied to our method to prevent the overfitting problem. This is achieved by grouping the factors in \eqref{eq1} by a set of two cliques $ C=\{C_u,C_b\} $, i.e., a unary one and a binary one. Then, we have

\begin{equation}
\label{eq3}
p(y|x;\theta)=\frac{1}{Z(x;\theta)}\prod_{C_F \in C}\prod_{f \in C_F}\text{e}^{-E_f(x_f,y_f;\theta_f)}.
\end{equation}
Defining $ E_f(x,y_f;\theta_f)=\theta_f\Phi_f(x,y_f)$, we have a more detailed form of \eqref{eq3} as follows
\begin{equation}
\label{eq4}
p(y|x;\theta)=\frac{1}{Z(x;\theta)}\text{e}^{\theta_u\sum\limits_{f\in C_u}\Phi_u(x_f,y_f)+\theta_b\sum\limits_{f\in C_b}\Phi_b(x_f,y_f)}.
\end{equation}
We can see that by calculating the energy functions of factors in two cliques, only two variables $ \theta_u $ and $ \theta_b $ are required to be optimized.

\begin{figure*}[!t]
	\centering{\includegraphics[width=0.8\textwidth]{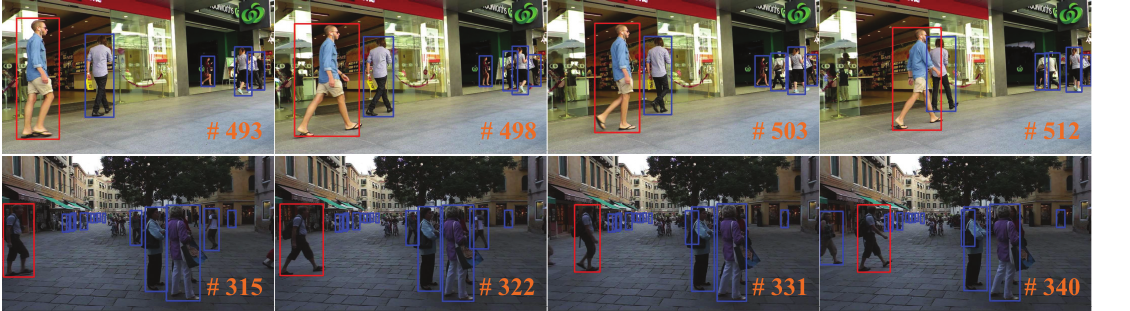}}
	\caption{Examples to explain for excluding the width term. The height of a pedestrian is stable regardless of the camera attitude, while the width of a pedestrian may change dramatically even in normal case because of the change in human postures.}
	\label{fig3}
\end{figure*}

\subsection{Feature Function Designing}
\label{subsec:feature_function_designing}
Following the above derivation, we now explicitly design the feature functions in the form of a unary term and a binary term. The unary term focuses on the reliability of individual tracklets, whereas the binary term considers the relationship between different pairs of tracklets.

\subsubsection{Unary Term}
\label{subsubsec:unary_term}
Two kinds of information are taken into account when designing the feature functions for the unary term: the classification score $ S_f $ and the changing rate of the aspect ratio $ \Delta R_f $:
\begin{equation}
\begin{aligned}
\label{eq5}
\Phi_u(x_f,y_f)=\textbf{1}_{\{y_f=0\}}(\mid 0-S_f\mid +\textbf{1}_{\{S_f>0.95\}}\alpha_1)+\\
\textbf{1}_{\{y_f=1\}}(\mid 1-S_f\mid+\alpha_2 \mid 1-\Delta R_f\mid),
\end{aligned}
\end{equation}
where
\begin{equation}
\label{eq6}
\textbf{1}_{\{A=B\}}= \left \{
\begin{array}{lr}
\displaystyle 1 \;   \ \  , \;  if \; A=B \\
0 \;   \ \  , \; if \; A \ne B
\end{array} \ \ .
\right.
\end{equation}
Here, $ \alpha_1 $ and $ \alpha_2 $ are hyperparameters that are initialized heuristically and are adjusted by using a validation dataset. $ S_f $ comes directly from the output of the Faster R-CNN. $ \textbf{1}_{\{S_f>0.95\}} $ lays extra punishment for the inactivation with classification scores higher than $ 0.95 $. To define $ \Delta R_f $ specifically, we replace it by $ \Delta R_i^t $, where $ i $ denotes the index of the only node in the factor. Then, we have

\begin{equation}
\label{eq7}
\Delta R_i^t=\frac{w_i^t/h_i^t}{w_i^{t-1}/h_i^{t-1}}.
\end{equation}

\subsubsection{Binary Term}
\label{subsubsec:binary_term}
In our method, the binary term formulates the punishment for joint values of the two nodes contained in each factor. In \cite{Zhou2019}, an assumption is given that the change of velocity should be basically the same between two tracking targets in a short time interval, i.e.,

\begin{equation}
\label{eq8}
disp_i^t-disp_j^t=disp_i^{t-1}-disp_j^{t-1},
\end{equation}
where $ disp_i^t$ is the center displacement of object $ i $ between frame $ t $ and frame $ t-1 $. This assumption has two implications:

(\romannumeral1) The absolute velocity of pedestrians can be approximately regarded as a constant value during short time intervals. Hence, $ disp_i^t-disp_i^{t-1} $ leads to the magnitude of camera motion.

(\romannumeral2) Sharp rotation and swing should not be involved in the camera motion.

We follow a similar assumption and make further improvement in the following four aspects:
\begin{enumerate}
  \item The units of measure is unified when calculating the velocity by multiplying the frame rate $ \omega $ of the sequence, i.e., $ V_i^t=\omega(P_i^t-P_i^{t-1}) $ where $ P $ denotes the center position of the object. Likely, we calculate the changing rate of velocity as $ \Delta V_i^t=\omega(V_i^t-V_i^{t-1}) $.
  \item A coefficient $ \tau=\frac{1}{h_i+h_j} $ is introduced to balance the weights, where $ h $ is the height of a tracked object. This design is proposed due to the observation that the position of a larger object tends to be estimated with worse performance.
  \item The above assumption is extended to the changing rate of the object height $ \Delta L_i^t= \omega\frac{\Delta h_i^t-\Delta h_i^{t-1}}{\Delta h_i^{t-1}}$, where $ \Delta h_i^t=\omega\frac{h_i^t-h_i^{t-1}}{h_i^{t-1}}$. Note that the width of object is cast aside when calculating $ \Delta L $, the reason for which is depicted in Fig.~\ref{fig3}.
  \item When part of a tacking target moves out of sight, $ \Delta L $ becomes unreliable due to the restriction of the image boundary. Therefore, an additional variable $ \kappa $ is introduced to address this concern, which takes $ 0 $ if the object is partially beyond the image boundary and takes $ 1 $ otherwise.
\end{enumerate}

Motivated by the above discussions, the feature functions of the binary term are designed as follows:
\begin{equation}
\begin{aligned}
\label{eq9}
\Phi_b(x_f,y_f)=\textbf{1}_{\{y_f=[1,1]\}} \bigl(\sum_{n\in\{x,y\}}\tau(\Delta V_{n,i}-\Delta V_{n,j})^2\\
+\beta \kappa\mid \Delta L_i-\Delta L_j \mid \; \bigr),
\end{aligned}
\end{equation}
where $\beta$ is a hyperparameter.

\subsection{Inference and Training}
\label{subsec:inference_and_training}
To facilitate the inference and training of a CRF, we leverage two techniques to keep the number of CRF nodes fixed. The first technique, termed as hypothesis filtering, filters out extra objects with the highest classification scores when the detected objects outnumber a predefined value $ M $. On the contrary, frames with nodes fewer than $ M $ are complemented by adding dummy nodes. The dummy nodes are artificially created to make the formulation hold, which are designed to exert no influence on the real nodes in a way that the formulae in Section \ref{sec:method} are regarded as multiplied by a boolean variable (which takes $ 1 $ when a factor contains no dummy nodes and takes $ 0 $ otherwise).

\subsubsection{CRF Inference}
\label{subsubsec:crf_inference}
The loopy belief propagation algorithm is adopted for dealing with the CRF inference task \cite{Kyburg1991}. The main idea of the loopy belief propagation algorithm is demonstrated as follows, and we refer interested readers to \cite{Kyburg1991,Nowozin2010,Sutton2011} for further mathematical details.

The message flowing from a factor node to a variable node is defined as
\begin{equation}
\label{eq10}
M_{f,v}(x,y_v;\theta)=\sum_{\substack{y_f \in Y_f(y_v)}}\bigl(\text{e}^{-E_f(x,y_f;\theta)}\prod\limits_{\substack{v^{\prime}\sim f\\ v^{\prime} \ne v}}M_{v^{\prime},f}(x,y_{v^{\prime}};\theta)\bigr),
\end{equation}
where $ v $ and $ f $ denote a variable node and a factor node, respectively. $ Y_f(y_u) $ is a subset of $ y_f $'s domain that requires the node $ u $ to take the value $ y_u $. $ v\sim f $ signifies that an edge exists between $ v $ and $ f $.

Conversely, the message flowing from a variable node to a factor node is defined as
\begin{equation}
\label{eq11}
M_{v,f}(x,y_v;\theta)=\dfrac{\prod\limits_{\substack{f^{\prime}\sim v \\ f^{\prime}\ne f }}M_{f^{\prime},v}(x,y_v;\theta)}{\sum\limits_{y_v \in Y_v}\prod\limits_{\substack{g\sim v \\ f^{\prime}\ne f }}M_{f^{\prime},v}(x,y_v;\theta)}.
\end{equation}

The marginal distribution is then calculated by

\begin{equation}
\label{eq12}
p(y_v|x;\theta)=\dfrac{\prod\limits_{f\sim v}M_{f,v}(x,y_v;\theta)}{\sum\limits_{y_v \in Y_v}\prod\limits_{f\sim u}M_{f,v}(x,y_v;\theta)}.
\end{equation}

In order to get the optimal output $ y^{\star} $ with respect to the maximum posterior probability (MAP), \eqref{eq10} and \eqref{eq11} are modified as
\begin{equation}
\label{eq13}
M_{f,v}(x,y_v;\theta)=\max_{\substack{y_f \in Y_f(y_v)}}\text{e}^{-E_f(x,y_f;\theta)}\prod\limits_{\substack{v^{\prime}\sim f\\ v^{\prime} \ne v}}M_{v^{\prime},f}(x,y_{v^{\prime}};\theta),
\end{equation}
\begin{equation}
\label{eq14}
M_{v,f}(x,y_v;\theta)=\frac{1}{2}\dfrac{\prod\limits_{\substack{f^{\prime}\sim v \\ f^{\prime}\ne f }}M_{f^{\prime},v}(x,y_v;\theta)}{\prod\limits_{y_v \in Y_v}\prod\limits_{\substack{g\sim v \\ f^{\prime}\ne f }}M_{f^{\prime},v}(x,y_v;\theta)}.
\end{equation}
Finally, the output labels are obtained as follows:
\begin{equation}
\label{eq15}
y^{\star}=[\;\max_{y_1} p(y_1|x;\theta),\dots,\max_{y_M} p(y_M|x;\theta)\;]^T.
\end{equation}

\subsubsection{CRF Training}
\label{subsubsec:crf_training}
The purpose of CRF training is to get an estimation of the parameters using training samples $ \{(x^1,y^1),\dots,(x^N,y^N)\} $. This problem can be naturally solved by Maximum Likelihood Estimation (MLE). Taking the first order derivative of the log likelihood of \eqref{eq4} and considering all $ N $ training samples, we have
\begin{equation}
\begin{aligned}
\label{eq16}
\frac{\partial l(\theta)}{\partial \theta_{u/b}}=\sum_{n=1}^{N}\bigl(\sum\limits_{f\in C_{u/b}}\Phi_{u/b}(x_f^n,y_f^n)-\\
\sum\limits_{f\in C_{u/b}}\sum\limits_{y_f^{\prime}}\Phi_{u/b}(x_f^n,y_f^{\prime})p(y_f^{\prime}|x^n;\theta)\;\bigr)
\end{aligned}
\end{equation}
where $ p(y_f|x;\theta) $ is calculated in a similar way to \eqref{eq12}:
\begin{equation}
\label{eq17}
p(y_f|x;\theta)=\dfrac{e^{-E_f(y_f)}\prod\limits_{v\sim f}M_{v,f}(x,y_v;\theta)}{\sum\limits_{y_f \in Y_f}\bigl[e^{-E_f(y_f)}\prod\limits_{v\sim f}M_{v,f}(x,y_v;\theta)\bigr]}.
\end{equation}

Applying the SGD algorithm with above gradients, the weights are updated with a fixed learning rate $ \gamma $:
\begin{equation}
\label{eq18}
\theta^{\prime}=\theta+\bigl[\;\gamma_u \frac{\partial l(\theta)}{\partial \theta_{u}}\;,\;\gamma_b \frac{\partial l(\theta)}{\partial \theta_{b}}\;\bigr]^T.
\end{equation}

\begin{algorithm}
	\caption{Workflow for tracking}
\label{alg1}
	\SetAlgoLined
	\KwIn{Images from frames to be tracked $ I^1,\dots,I^T $,
		Public detection results $ D^1,\dots,D^t $.}
	\KwOut{Tracklets of objects $ tr_1,...,tr_K $.}
	Initialize tracklets $ tr_1,...,tr_K $ by detections $ D^1 $;\\
	\For{$t=2,\dots,T$}
	{
		Obtain regressing results $ x^t $ and classification scores $ S^t $ by using Tracktor++;\\
		\For{tracklet index $ i=1,\dots,K^t $}
		{
			\eIf{$ S_i^t< 0.4$}
			{
				Inactivate tracklet $ i $;\\
			}{
				\If{tracklet length $ C_i \ge 2$}
				{
					Calculate all the tracking information required in \ref{subsec:feature_function_designing};\\
				}
			}
		}
		\If{number of tracklets $ K^t > 10 $ }
		{Cut down $ K^t $ to $ 10 $ by hypothesis filtering (\ref{subsec:inference_and_training});
		}
		Calculate the unary and binary terms in the CRF (\ref{subsec:feature_function_designing});\\
		Obtain the inactivation labels by using the loopy belief propagation algorithm (\ref{subsubsec:crf_inference});\\
		Inactivate tracklets by using the labels;\\
		Do non-maximum suppression and reidentification.\\
		Update $ tr_1,...,tr_k$;
	}
	\Return $ tr_1,...,tr_k$.
\end{algorithm}

\section{Implementation details}
\label{sec:implementation_details}
In this section, we first introduce the datasets used in practical tracking. The details for parameter estimation are then presented. Finally, the workflow and other details of practical tracking are introduced.

\subsection{Datasets}
\label{subsec:datasets}
Two popular datasets, MOT16 and MOT17 \cite{Milan2016}, are employed in this paper. Both these datasets provide official detection results that are called public detections. Researchers can also use private detections produced by their own detectors.
\subsubsection{MOT16}
The MOT16 dataset contains 14 video sequences of pedestrians with different resolutions, frame rates, lighting conditions, crowd density, and filming angles. They are evenly divided into a training dataset and a testing dataset. Each dataset has 4 sequences with moving cameras and 3 sequences with static cameras. The public detections are produced by using a DPM detector \cite{Felzenszwalb2010a}.
\subsubsection{MOT17}
The MOT17 dataset has the same video sequences as the MOT16. Notice that MOT17 provides three groups of public detections, which are produced by Faster R-CNN \cite{Ren2017a}, DPM \cite{Felzenszwalb2010a}, and SDP \cite{Yang2016a} respectively, resulting in 42 sequences altogether. Trackers' performance on all 21 test sequences are averaged to get a generalized evaluation result. When tracking is performed with private detections, there is no difference in using the MOT16 dataset or the MOT17 dataset.

\begin{table}[htbp]
	\centering
	\caption{Parameters in CRF}
	\label{tab1}
	\resizebox{0.35\textwidth}{!}{%
		\begin{tabular}{cccccc}
			\toprule
			Parameter
			&$ \theta_u $&$ \theta_b $&$ \alpha_1 $&$ \alpha_2 $&$ \beta $\\
			\midrule
			Value
			&0.98&0.12&1.05&1.20&10.80\\
			
			\bottomrule	
		\end{tabular}%
	}
\end{table}

\begin{table}[htbp]
	\centering
	\caption{Comparison of running speed (Hz)}
	\label{tab2}
	\resizebox{0.44\textwidth}{!}{%
		\begin{tabular}{cccccc}
			\toprule
			Hypothesis number
			& 3 & 6 & 10 & 15&Overall\\
			\midrule
			Ours
			&6.80&5.51&2.82&1.67&2.22\\

			Tracktor++
			&7.09&6.35&3.20&1.98&2.58\\
			\bottomrule
		\end{tabular}%
	}
\end{table}

\begin{table*}[htbp]
	\centering
	\caption{MOT16 results (public detections)}
	\label{tab3}
	\resizebox{0.75\textwidth}{!}{%
		\begin{tabular}{cccccccccc}
			\toprule
			&MOTA$\uparrow$&IDF1$\uparrow$&MOTP$\uparrow$&MT$\uparrow$&ML$\downarrow$
			&FP$\downarrow$&FN$\downarrow$&IDS$\downarrow$&Frag$\downarrow$ \\
			\midrule
			Ours
			&\textbf{57.0}&\textbf{58.3}&79.1&\textbf{158}&\textbf{267}&2610&\textbf{75337}&538&1212 \\
			
			Tracktor++v2 \cite{Bergmann2019}
			&56.2&54.9&\textbf{79.2}&157&272&\textbf{2394}&76844&617&1068 \\
			
			KCF16\cite{Chu2019a}
			&48.8&47.2&75.7&120&289&5875&86567&906&1116 \\
			
			MOTDT\cite{Chen2018a}
			&47.6&50.9&74.8&115&291&9253&85431&792&1858 \\
			
			JCSTD\cite{Tian2020}
			&47.4&41.1&74.4&109&276&8076&86638&1266&2697 \\
			
			AMIR\cite{Sadeghian2017a}
			&47.2&46.3&75.8&106&316&2681&92856&\textbf{370}&\textbf{598} \\
			
			YOONKJ\cite{Yoon2020}
			&47.0&50.1&75.8&125&317&7901&88179&627&945 \\
			
			DD\_TAMA\cite{Yoon2019}
			&46.2&49.4&75.4&107&334&5126&92367&598&1127 \\
			
			DMAN\cite{Zhu2018b}
			&46.1&54.8&73.8&132&324&7909&89874&532&1616 \\
			
			STAM\cite{Chu2017b}
			&46.0&50.0&74.9&111&331&6895&91117&473&1422 \\
			
			RAR16pub\cite{Fang2018}
			&45.9&48.8&74.8&100&318&6871&91173&648&1992 \\
			\bottomrule	
		\end{tabular}%
	}
\end{table*}

\begin{table*}[htbp]
	\centering
	\caption{MOT17 results (public detections)}
	\label{tab4}
	\resizebox{0.75\textwidth}{!}{%
		\begin{tabular}{cccccccccc}
			\toprule
			&MOTA$\uparrow$&IDF1$\uparrow$&MOTP$\uparrow$&MT$\uparrow$&ML$\downarrow$
			&FP$\downarrow$&FN$\downarrow$&IDS$\downarrow$&Frag$\downarrow$ \\
			\midrule
			Ours
			&\textbf{56.3}&57.5&\textbf{78.9}
			&490&839&\textbf{8672}&236296&\textbf{1668}&3820 \\
			
			Tracktor++v2 \cite{Bergmann2019}
			&\textbf{56.3}&55.1&78.8&498&831&8866&\textbf{235449}&1987&3763 \\
			
			LSST\cite{Feng2019}
			&52.7&\textbf{57.9}&76.2&421&801&15884&246939&3711&8757 \\
			
			FAMNet\cite{Chu2019b}
			&52.0&48.7&76.5&450&787&14138&253616&3072&5318 \\
			
			YOONKJ\cite{Yoon2020}
			&51.4&54.0&77.0&500&878&29051&243202&2118&\textbf{3072} \\
			
			STRN\cite{Xu2019}
			&50.9&56.0&75.6&446&797&25295&249365&2397&9363 \\
			
			MOTDT\cite{Chen2018a}
			&50.9&52.7&76.6&413&841&24069&250768&2474&5317 \\
			
			DEEP\_TAMA\cite{Yoon2019}
			&50.3&53.5&76.7&453&883&25479&252996&2192&3978 \\
			
			EDMT\cite{Chen2017a}
			&50.0&51.3&77.3&\textbf{509}&855&32279&247297&2264&3260 \\
			
			GMPHDOGM\cite{Song2019}
			&49.9&47.1&77.0&464&895&24024&255277&3125&3540 \\
			
			MTDF\cite{Fu2019}
			&49.6&45.2&75.5&444&\textbf{779}&37124&241768&5567&9260 \\
			\bottomrule	
		\end{tabular}%
	}
\end{table*}

\begin{table*}[htbp]
	\centering
	\caption{MOT17 results (private detections)}
	\label{tab5}
	\resizebox{0.75\textwidth}{!}{%
		\begin{tabular}{cccccccccc}
			\toprule
			&MOTA$\uparrow$&IDF1$\uparrow$&MOTP$\uparrow$&MT$\uparrow$&ML$\downarrow$
			&FP$\downarrow$&FN$\downarrow$&IDS$\downarrow$&Frag$\downarrow$ \\
			\midrule
			Ours
			&58.9&60.4&78.1
			&609&669&20565&208680&2544&6459 \\
			
			Tracktor++v2
			&58.9&56.9&78.1&603&666&20640&208572&2853&6471 \\
			
			\bottomrule	
		\end{tabular}%
	}
\end{table*}

\subsection{Paramater Estimation}
\label{subsec:paramater_estimation}
It should be mentioned that there is no suitable existing dataset for training our model. In this case, a specific dataset is created on our own to facilitate the training of our CRF module. The training datasets of MOT16 and MOT17 are divided into two parts: one part with $ 40\% $ video frames is used to generate the training dataset, and the other one serves as the validation dataset. The original Tracktor++ is applied to the first part and pick out the tracking errors produced by improper inactivation for tracklets. Frames with such errors are treated as negative samples. Note that $ 3 $ times more positive samples are randomly selected. The weights of the unary and binary terms are then trained by using the SGD algorithm with a learning rate of $ 1\times 10^{-2} $ for $ 30 $ epochs.

Experimental results have shown that directly using the trained weights gave rise to unpromising performance on the validation dataset. This phenomenon stems from two sources:

(\romannumeral1) Tracktor++ itself is a powerful tracker so that the generated training dataset has limited capacity.

(\romannumeral2) The samples from the dataset do not strictly satisfy the premise of independent and identical distribution (i.i.d.).

Thus, we use the weights learned in the training phase as a starting point wherefrom heuristic adjustment is made. Here we first apply the initial values to the hyperparameters and manually select some obvious failing cases from the tracking results on the training dataset. We observe the value of each item in \eqref{eq5} and \eqref{eq6}. Then $ \alpha_1 $ is adjusted to balance the values between normal classification scores and high ($>0.95$) classification scores. $ \alpha_2 $ is adjusted to balance the values between classifications scores and the changing of aspect ratios. $ \beta $ is adjusted to balance the values between the changing rate of velocity and object height. Eventually, $ \theta_u $ and $ \theta_u $ are adjusted so that neither the unary term nor the binary term should be too much larger than each other. In fact, $ \alpha_1 $, $ \alpha_2 $ and $ \beta $ can also be trained by using the SGD algorithm instead of being treated as hyperparameters. The final values for all the parameters in our method are listed in Table~\ref{tab2}.

\subsection{Practical Tracking}
\label{subsec:practical_tracking}
The overall workflow for practical tracking with public detections is shown in Algorithm \ref{alg1}. Before performing the hypothesis filtering, all the tracklets with classification scores lower than $ 0.4 $ are inactivated. As for the predefined value $ M $ in \ref{subsec:inference_and_training}, we need to seek a decent balance between the handling capability of the CRF and the tracking speed. Notice that it is scarcely possible for a tracker to simultaneously inactivate more than $ 5 $ objects in a single frame. In this case, we double the number and set $ M=10 $, meaning that $ 10 $ tracking hypotheses with the lowest classification scores are modelled as CRF nodes. Frames with objects less than $10$ will be processed with dummy nodes as mentioned. For tracklets whose lengths are shorter than 3, we follow the convention in Tracktor++, i.e., a threshold (0.5) of classification score is selected to determine the inactivation.

To present a fair comparison with the baseline tracker, we wrote the experimental code based on the newly released code\footnote{https://github.com/phil-bergmann/tracking\_wo\_bnw} provided by the authors of Tracktor++. Because of a retrained CNN network, the performance of the released code surpasses that of the original paper. When comparing the experimental tracking performance in the following section, we refer to this version of tracker as Tracktor++v2, which is an alias used by its authors on the MOT16 and MOT17 results.

The computational complexity of the CRF module is $\mathcal{O}(n^2)$ in view of the number of CRF nodes. We test our tracker and the released code of Tracktor++ on a personal computer with an Intel i7-9700 CPU and a Nvidia GTX 2080ti GPU. The average running speed of both trackers with respect to sequence frames containing different number of tracking hypotheses is reported in Table~\ref{tab1}.

\section{Experiments}
\label{sec:experiments}
In this section, we first introduce the metrics used to evaluate the MOT performance. Then, quantitative results on MOT16 and MOT17 benchmarks are presented. After that, the tracking results are shown qualitatively to provide an intuitive understanding of our method. Finally, an extensive experiment is conducted to exhibit the extensibility of our method.

\subsection{Evaluation Metrics}
\label{subsec:evaluation_metrics}
Evaluating the performance of an MOT tracker reasonably is a non-trivial task. Three evaluation systems, the CLEAR MOT metrics \cite{Bernardin2008}, the ID metrics \cite{Ristani2016} and some other classical metrics are used to verify the validity of our method. These metrics are regrouped into two categories as follows:

(\romannumeral1) Singular metrics.
\begin{itemize}
	\item Multiple Object Tracking Precision (MOTP). MOTP describes the precision of bounding box regression, which mainly relies on the performance of the object detector.
	\item Mostly Tracked (MT). The number of trajectories that are successfully tracked for more than 80\% of its ground truth boxes.
	\item Mostly Lost (ML). The number of trajectories that fail to be tracked for more than 20\% of its ground truth boxes.
	\item False Positives (FP).
	\item False Negatives (FN).
	\item Identification Switches (IDS).
	\item Fragments (Frag). The number of fragments in all trajectories.
\end{itemize}

(\romannumeral2) Compound metrics.
\begin{itemize}
	\item
Multiple Object Tracking Accuracy (MOTA). Considering the number of False Positives (FP), False Negatives (FN), and ID Switches (IDS), MOTA is defined as
\begin{equation}
\label{eq19}
\text{MOTA}=1-\frac{\text{FP}+\text{FN}+\text{IDS}}{\text{GT}},
\end{equation}
where GT denotes the number of ground truth boxes.

\item
Identification $ \text{F}_1 $ (IDF1). IDF1 stands for the $ \text{F}_1 $ score of ID Precision (IDP) and ID Recall (IDR):
\begin{equation}
\label{eq20}
\text{IDF1}=1-\frac{2}{\frac{1}{IDP}+\frac{1}{IDR}}=\frac{2\text{IDTP}}{2\text{IDTP}+\text{IDFP}+\text{IDFN}},
\end{equation}
where IDTP, IDFP, and IDFN stands for ID True Positives, ID False Positives, and ID False Negatives respectively. Detailed definitions can be found in \cite{Ristani2016}.
\end{itemize}
\begin{figure}[h]
	\centering
	\includegraphics[width=0.33\textwidth]{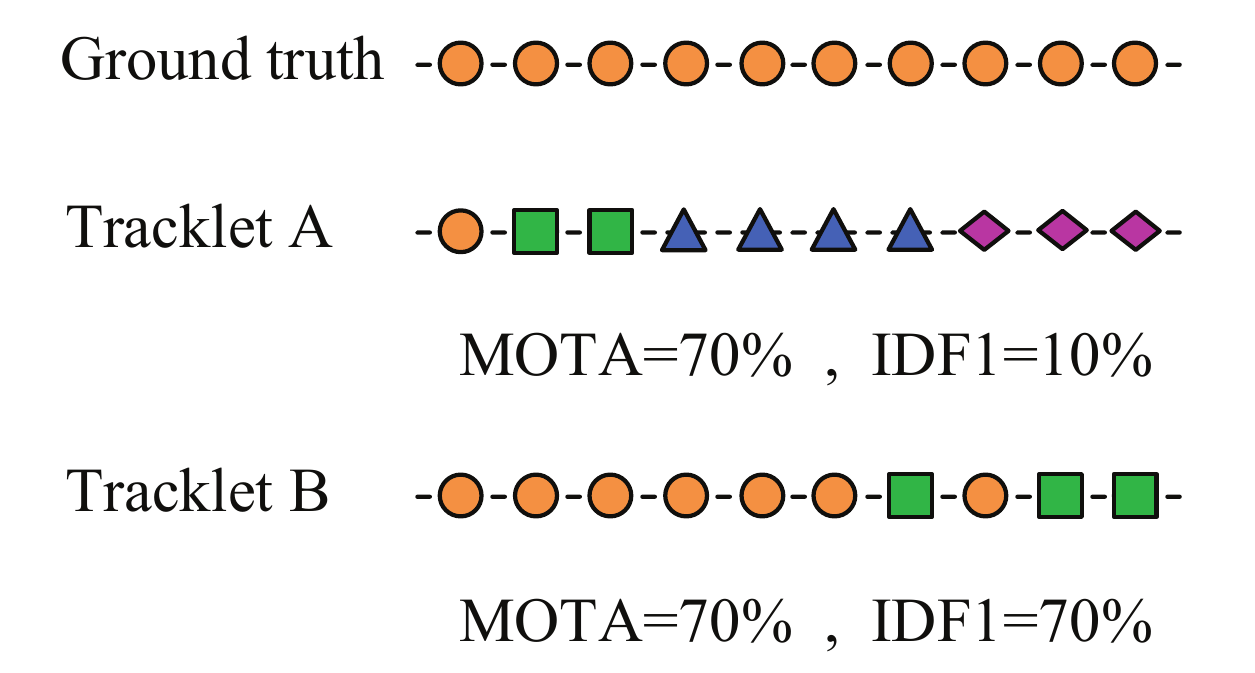}
	\caption{Illustration for the deficiency to use MOTA alone. Tracklet A and tracklet B have the same MOTA score, yet the latter one is intuitively better. This is because the CLEAR MOT metrics pay more attention to the coverage rate of the ground truth boxes, while ID metrics are able to depict more details about the ID information.}
	\label{fig4}
\end{figure}

The compound metrics are chosen as the main indicators to evaluate trackers' performance. Although MOTA is a comprehensive metric, it is not enough to judge a trackers' performance only depending on MOTA. We illustrate the reason why MOTA should be combined with IDF1 in Fig.~\ref{fig4}. The assumption is that all the objects are perfectly recognized whereas the identifications are improperly judged. Different assumed identifications are represented by different shapes and colors.

\subsection{Quantitative Analysis}
\label{subsec:quantitative_analysis}
We apply the proposed framework to Tracktor++ and compare the resultant tracker with TOP-10 online MOT trackers\footnote{We only consider the results on the MOT official website with published works. The TOP-10 list is up to the date when this paper was written.} with public detections on the MOT16 and MOT17 benchmarks. The experimental results are listed in Table~\ref{tab3} and Table~\ref{tab4}. The up-arrow denotes the higher the better and the down-arrow means the opposite. The highest scores are marked in bold.

Taking MOTA and IDF1 scores as the first and second important metrics, our method achieves the best performance on both MOT16 and MOT17 benchmarks. Compared with the baseline method Tracktor++ on MOT16, the MOTA and IDF1 scores are increased by $ 0.8\% $ and $ 3.4\% $, respectively. On MOT17, we increase the IDF1 score by $ 2.4\% $. In Table~\ref{tab4}, the IDF1 score of our method performs better than the others' in evaluating a tracker's ability to maintain the identification of tracklets. This result is coherent with our improved approach for tracklet inactivation, which mitigates the problem of hypothesis drifting.

We further compare our tracker with Tracktor++ using private detections on the MOT17 benchmark, i.e., objects are detected by employing the retrained Faster R-CNN of both our tracker and Tracktor++. Such comparison is not done on MOT16 since the sequences in MOT16 are identical to those in MOT17. As is shown in Table~\ref{tab5}, we increase the IDF1 score by $ 3.5\% $. It is worth mentioning that the ID Switches (IDS) of our approach is less than that of the Tracktor++ in all three experimental setups, which further confirms that our method is important in stabilizing the ID information.

\begin{figure}[htbp]
	\centering{\includegraphics[width=0.45\textwidth]{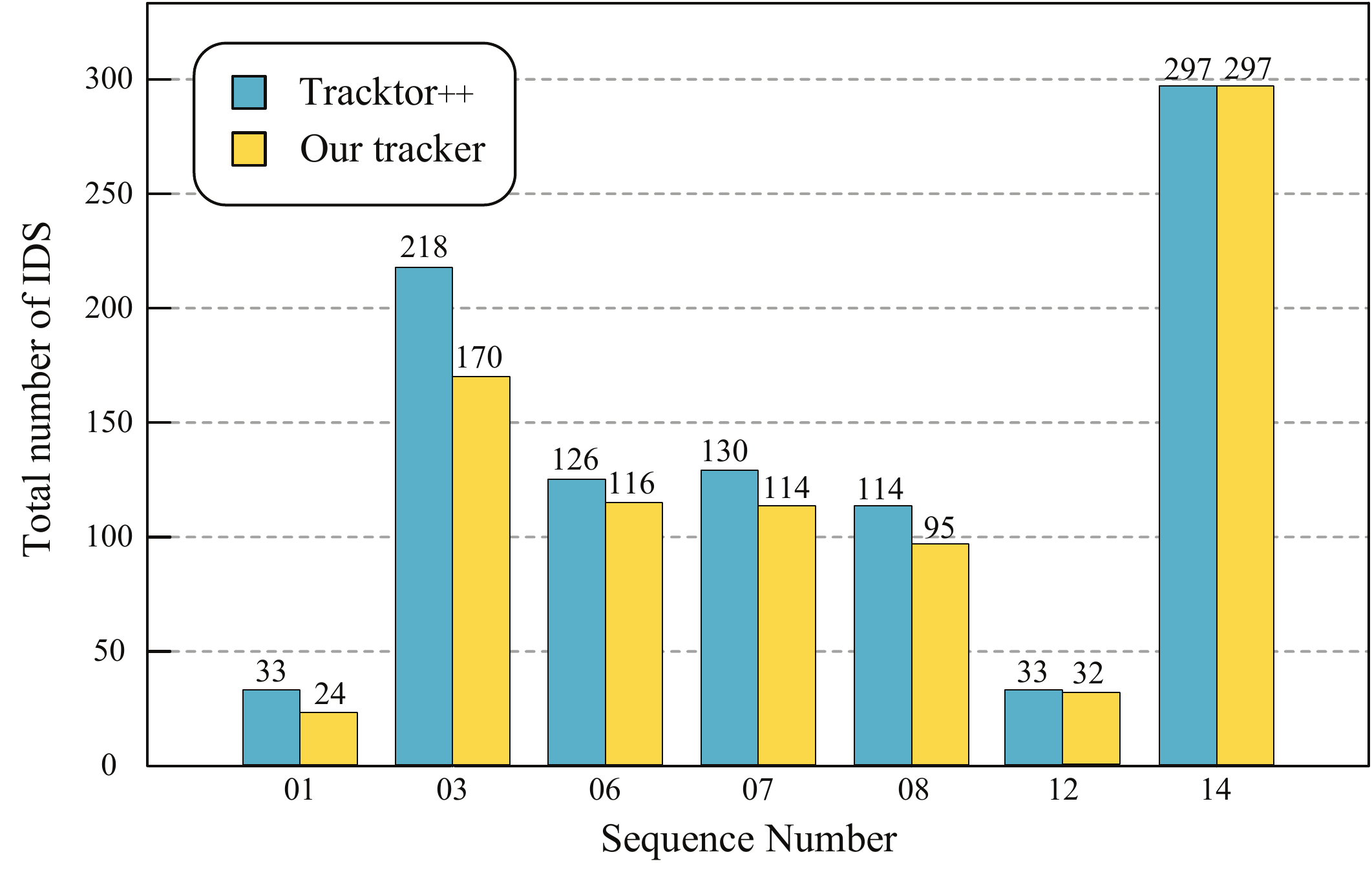}}
	\caption{Statistics of IDS in MOT17 test sequences (with private detections). }
	\label{fig5}
\end{figure}

For a deeper insight of the experimental results, we count the number of IDS in each one of MOT17 test sequences, as is shown in Fig.~\ref{fig5}. We see that our method alleviates the problem of ID Switch in both sequences with a static camera (01, 03, 08) and sequences with camera motion (06, 07, 12). The only test result with no IDS reduction is MOT17-14, the scenes in which rotate a lot and have a sharp left and right swing. In such a case, our assumption in subsection \ref{subsec:feature_function_designing} does not hold well as in other video sequences.

\begin{figure}[htbp]
	\centering
	\subfloat[MOT16]{\includegraphics[width=0.45\textwidth]{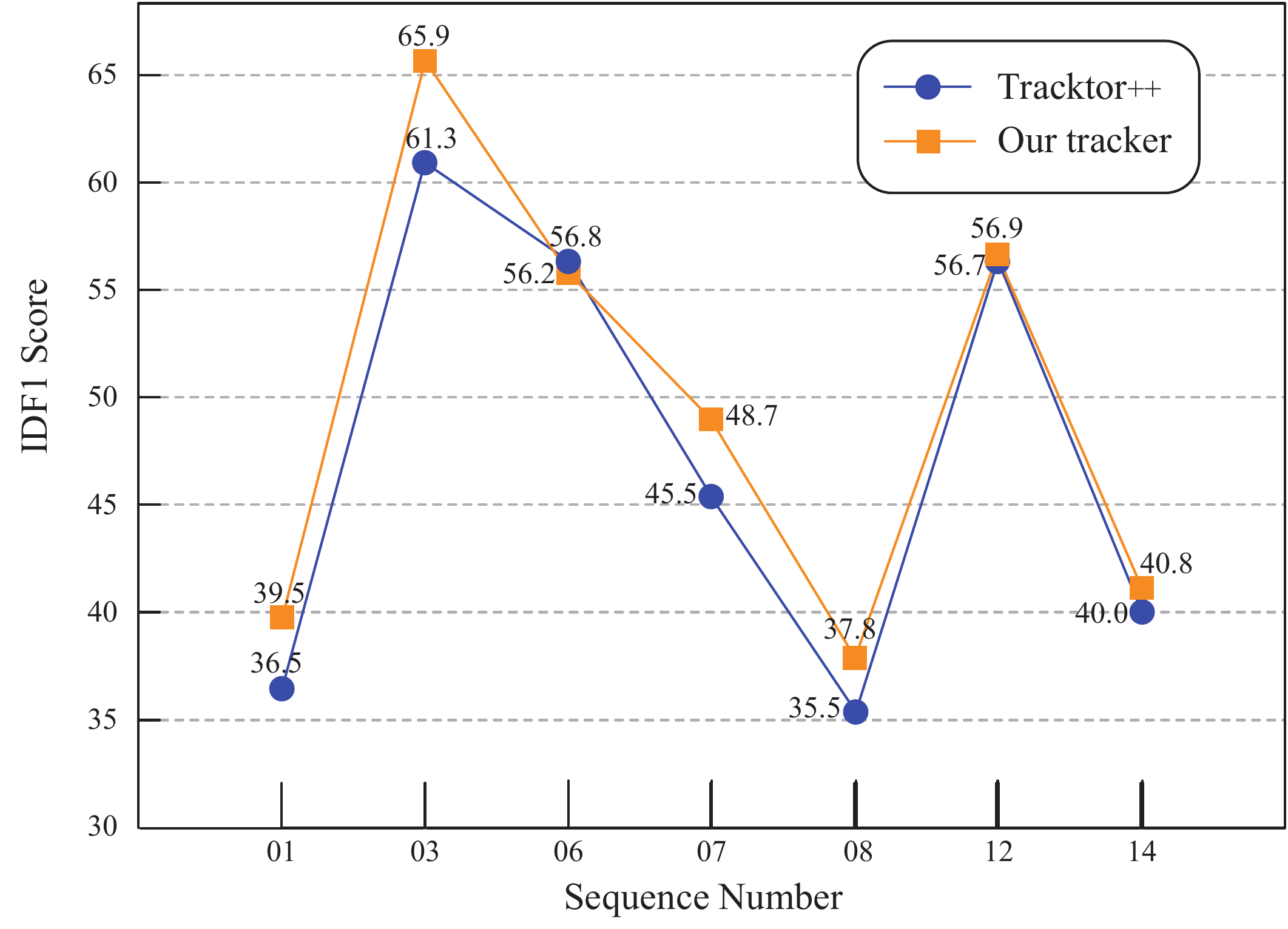}}\quad
	\subfloat[MOT17]{\includegraphics[width=0.45\textwidth]{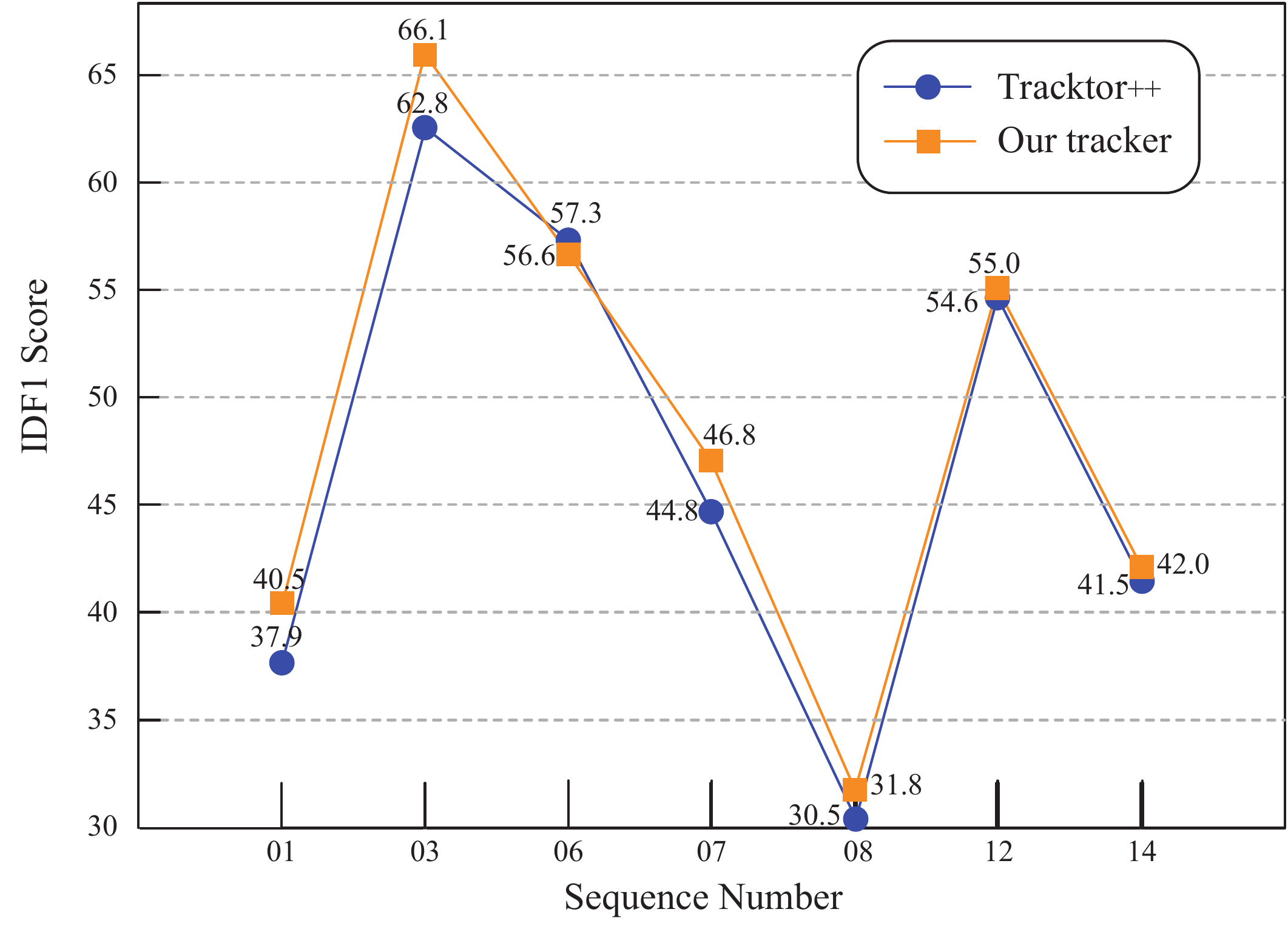}}\quad 
	\caption{Comparision of IDF1 scores on MOT benchmarks.}
	\label{fig6}
\end{figure}

We also make a comparison between the baseline tracker and our improved tracker in view of a significant MOT metric, the IDF1 score, in Fig.~\ref{fig6}. Compared with IDS, the IDF1 score can better reflect the continuity and uniqueness of ID assignment and thus is an investigation focus of our method. It can be observed that our tracker achieves a higher IDF1 score than the baseline tracker in most test sequences on both MOT16 and MOT17 benchmarks. Note that our tracker is not superior on all the video sequences because of the domain diversity of MOT benchmark. The video sequences in MOT are captured under various environmental settings, such as different illumination conditions, different camera angles, with or without camera motion, etc. This means that we have to make a compromise on the parameters to achieve a generally decent performance for multiple domains. However, in real applications (e.g. intelligent surveillance), we scarcely need a tracker to cope with data from as many domains as in the MOT challenge. In other words, our method has more potential to be exploited than it seems on the benchmark performance.

\begin{figure*}[htbp]
	\centering{\includegraphics[width=0.9\textwidth]{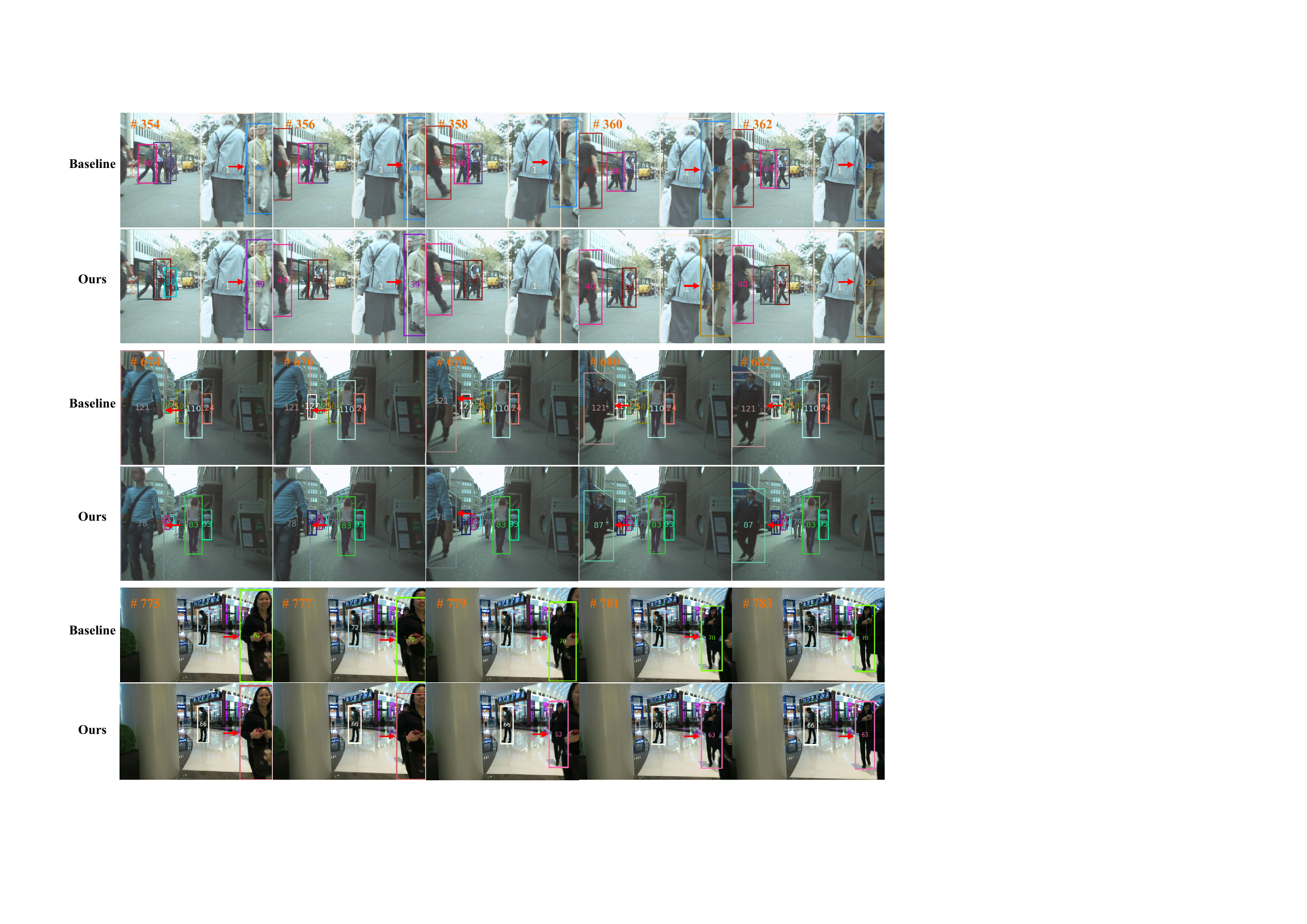}}
	\caption{Some visualized tracking results. We pick video clips from MOT17-05, MOT16-06, and MOT16-07, all of which are video sequences with camera motion. The 1st, 3rd, and 5th rows are the tracking results by Tracktor++, while the 2nd, 4th, and 6th rows are our results. By observing the target indicated by the red arrow, we see that drifting could happen when occlusion occurs. Using a discrete CRF, we resolve such problems by reasonable judgement about when to inactivate a tracklet.}
	\label{fig7}
\end{figure*}

\begin{figure*}[htbp]
	\centering{\includegraphics[width=0.9\textwidth]{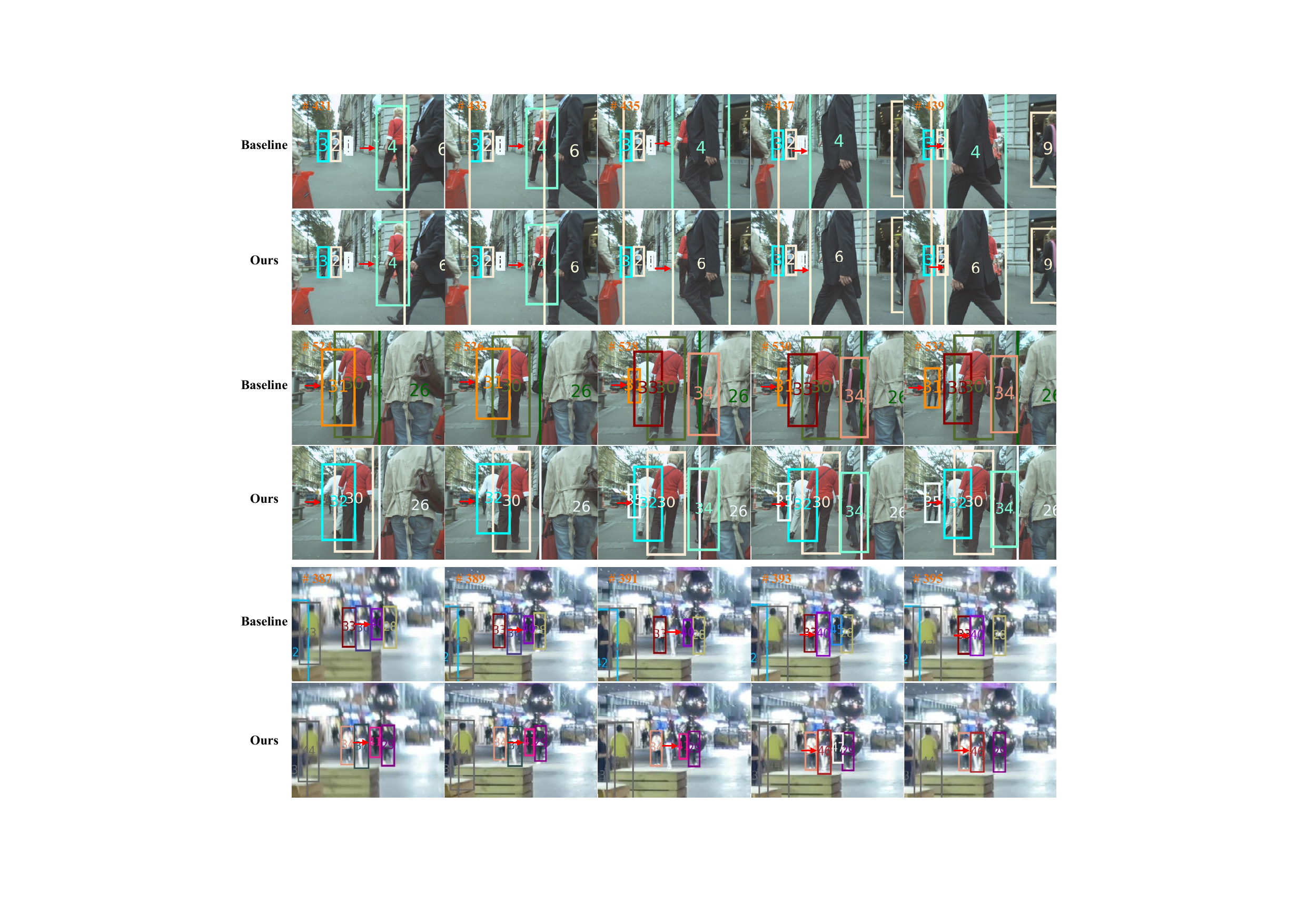}}
	\caption{Some visualized tracking results of the extensive experiment. The first two video clips come from MOT17-05 the third one comes from MOT17-10. The 1st, 3rd, and 5th rows are the tracking results by CenterTrack, while the 2nd, 4th, and 6th rows are our results. The tracking targets to be noticed are indicated by red arrows.}
	\label{fig8}
\end{figure*}

\subsection{Qualitative Analysis}
\label{subsec:qualitative_analysis}
We visualize some of the representative tracking results (with public detections) in Fig.~\ref{fig7} to provide an intuitive understanding of our method. The tracking hypotheses to be noticed are marked with red arrows. For narrative convenience the hypotheses pointed by red arrows are dubbed as target people. In the 1st row, when the target person is leaving the image boundary at frame 358, its bounding box begins to drift to the man in black shirt. The tracker mistakes the fake target person as the original one and continues to track him without changing the tracking ID in later frames. By contrast, our improved tracker correctly inactivates the target person at frame 358. The newly emerging person in black shirt is then treated as a new tracking object so that a new tracking ID is assigned to him. Likewise, the target person in the 3rd row drifts to the man in suit at frame 680 and the target person in the 5th row drifts to the nearby woman at frame 779. It can be seen from the 4th and the 6th rows that our method kills such drifting trend by inactivating the tracklets in due course.

When an occluded object gradually enters the visual field, it is supposed to be either identified as a new track by the detector or reidentified as a previous track by a CNN network. However, neither of these circumstances would be accessible when the object is compulsively assigned a wrong ID due to nearby hypothesis drifting. Even worse, the tracker is unable to be aware of such an ID error since new position is regressed merely using the position in the previous frame. Our CRF model handles this problem by telling the tracker to inactivate the tracklet promptly before drifting happens, making possible new detection and reidentification that would be suppressed otherwise.

It may be argued about the necessity to consider the inter-object relationship as we do, because hypothesis drifting can be recognized by discovering the sudden change of the position or size of a hypothesis. This argument does not hold when there is camera motion in the video sequences. On the contrary, our algorithm still works out in the existence of camera motion.

\begin{table}[htbp]
	\centering
	\caption{Ablation study on CRF parameters with the metric IDF1}
	\label{tab6}
	\resizebox{0.48\textwidth}{!}{%
		\begin{tabular}{cccccccc}
			\toprule
			Parameters&-50\%&-20\%&-10\%&ref&+10\% &+20\%&+50\%\\ \midrule
			$\theta_u$&66.3&66.7&66.8&66.9&66.9&66.8&66.4\\
			$\theta_b$&66.4&66.6&66.8&66.9&66.8&66.7&66.4\\ 
			$\alpha_1$&66.5&66.6&66.8&66.9&66.8&66.7&66.6\\
			$\alpha_2$&66.5&66.7&66.9&66.9&66.8&66.7&66.5\\
			$\beta$&66.5&66.6&66.8&66.9&66.9&66.8&66.6\\ \bottomrule
		\end{tabular}%
	}
\end{table}

\begin{table}[htbp]
	\centering
	\caption{Ablation study on CRF parameters with the metric IDS}
	\label{tab7}
	\resizebox{0.48\textwidth}{!}{%
		\begin{tabular}{cccccccc}
			\toprule
			Parameters&-50\%&-20\%&-10\%&ref&+10\% &+20\%&+50\%\\ \midrule
			$\theta_u$&247&238&234&232&233&236&243\\
			$\theta_b$&244&236&232&232&233&237&244\\ 
			$\alpha_1$&244&235&234&232&234&237&239\\
			$\alpha_2$&242&238&233&232&234&238&242\\
			$\beta$&241&237&234&232&233&236&240\\ \bottomrule
		\end{tabular}%
	}
\end{table}

\subsection{Ablation Study}
To investigate the influence on tracking performance posed by parameter variation, the parametric values in Table~\ref{tab1} are taken as the reference (denoted by `ref') and varied in different degrees for ablation study. We focus on the IDF1 and IDS metrics since they are most directly related to the proposed framework. The experimental results are reported in Table~\ref{tab6} and Table~\ref{tab7}, respectively. All the experiments are under controlled setups, i.e., when one parameter changes the remaining parameters are kept unchanged. The MOT17 training dataset with the Faster R-CNN split of public detections is used for evaluation.

Specifically, $\theta_u$ and $\theta_b$ account for the weights of the unary and binary terms. By increasing $\theta_b$, a larger proportion of the final inactivation labels will be determined by inter-object relationships. Likewise, a larger $\theta_u$ lays more emphasis on the behaviors of individual tracking hypotheses. $\alpha_1$, $\alpha_2$ and $\beta$ are employed to balance the contributions of different items in the feature functions in (\ref{eq5}) and (\ref{eq9}). Thus, none of the parameters $\{\theta_u, \theta_b, \alpha_1, \alpha_2, \beta\}$ should be either too large or too small, so that the whole system will not be dominated by any single constituent. This explains the performance drop (lower IDF1 and more ID switches) induced by the strong variation (at $\pm 50\%$) in parametric values. On the other hand, it can be observed that the overall performance is relatively insensitive to parameter variation within the range of $\pm 10\%$, indicating that the proposed framework has decent robustness to parameter variation.

\begin{table*}[htbp]
	\centering
	\caption{MOT17 results of the extensive experiment (public detections)}
	\label{tab8}
	\resizebox{0.95\textwidth}{!}{%
		\begin{tabular}{ccccccccccccccc}
			\toprule
			\multirow{2}{*}{Method} & \multicolumn{2}{c}{Seq. 05} & \multicolumn{2}{c}{Seq. 10} & \multicolumn{2}{c}{Seq. 11}& \multicolumn{2}{c}{Seq. 02}& \multicolumn{2}{c}{Seq. 04}& \multicolumn{2}{c}{Seq. 09}& \multicolumn{2}{c}{Seq. 13} \\ 
			&IDF1$\uparrow$&IDS$\downarrow$&IDF1$\uparrow$&IDS$\downarrow$&IDF1$\uparrow$&IDS$\downarrow$&IDF1$\uparrow$&IDS$\downarrow$&IDF1$\uparrow$&IDS$\downarrow$&IDF1$\uparrow$&IDS$\downarrow$&IDF1$\uparrow$&IDS$\downarrow$\\ \midrule
			Ours&61.4&49&56.8&76&59.6&27&37.1&97&75.6&103&60.7&38&62.0&72\\
			CenterTrack&60.2&52&55.8&78&58.8&28&37.0&98&75.4&110&60.7&39&61.9&74\\ \bottomrule
		\end{tabular}%
	}
\end{table*}

\subsection{Extensive Experiment}
\label{subsec:Extensive Experiment}
Due to the fact that the proposed method is decoupled from the main tracking pipeline of Tracktor++, we can adapt it to other trackers that are short of robust tracklet inactivation mechanism. To give an example, we conduct an extensive experiment on another state-of-the-art method CenterTrack \cite{zhou2020tracking} to exhibit the extensibility and transferability of our method.

CenterTrack takes both the current frame and the previous frame as input and simultaneously predicts the object locations and the inter-frame offsets. A greedy matching algorithm is utilized to associate the detections and tracklets. To apply our method, we consider the inter-object relationship and remove the associated pairs that are denied by the CRF module. Then we re-associate the remaining unassigned detections and tracklets by using the greedy matching algorithm.

The unary term of the feature function is adapted into:
\begin{equation}
\begin{aligned}
\label{eq21}
\Phi_u(x_f,y_f)=\textbf{1}_{\{y_f=0\}}\frac{1}{D}+\textbf{1}_{\{y_f=1\}}\alpha \mid 1-\Delta R_f\mid),
\end{aligned}
\end{equation}
where $D$ is the calculated distance between the center of the tracked position and the biased center (i.e., taking into account the predicted offset) of the detection. The binary term of the feature function is kept unchanged.

As is done in \cite{zhou2020tracking}, the training dataset of MOT17 is split into a training set and a validation set. We tune the parameters only on the training set and evaluate the performance of our improved method on the validation set. The experimental results are reported in Table~\ref{tab8}. Compared with the baseline, the tracker refined by our framework is able to achieve a higher IDF1 score and yield fewer ID switches, revealing the extensibility and applying prospect of the proposed method. It should be noted that our framework enhances the baseline tracker by dealing with tracklet inactivation, the effectiveness of which is mainly reflected on the reduction of ID related mistakes. By inspection, it is found that the baseline tracker makes very few such mistakes with the video sequences 02, 04, 09 and 11. This explains the reason why the baseline tracker benefits less obviously from the proposed framework with the last four sequences in Table~\ref{tab8}.
 
For an intuitive understanding of our method, some of the visualized results are shown in Fig.~\ref{fig8}. With the consideration of inter-object relationship, the CRF module contributes to killing the wrong ID assignment by detecting the abnormal changes of the tracking hypotheses in a tracklet. This can be seen by observing the behavioral differences between our tracker and CenterTrack at frames 433 $\rightarrow$ 435, frames 526 $\rightarrow$ 528, and frames 391 $\rightarrow$ 393 respectively in the three video clips of Fig.~\ref{fig8}.

\section{Conclusion}
\label{sec:conclusion}
In this paper, a CRF-based framework has been put forward to address the problem of unreliable tracklet inactivation in the context of online multi-object tracking. The behaviors of individual tracking hypotheses and inter-object relationships are collectively modelled by a discrete conditional random field. Dedicated feature functions have been designed to cope with various challenges in practical tracking scenarios. To handle the problem of varying CRF nodes, two simple yet effective strategies have been proposed. By applying the framework to a state-of-the-art MOT tracker, the resultant refined tracker has proven to outperform the baseline through experiments on the MOT16 and MOT17 datasets. The extensibility of the proposed method has been further corroborated by an extensive experiment.

\bibliographystyle{IEEEtran}
\bibliography{paper_2020}

\begin{IEEEbiography}[{\includegraphics[width=1in,height=1.25in,clip,keepaspectratio]{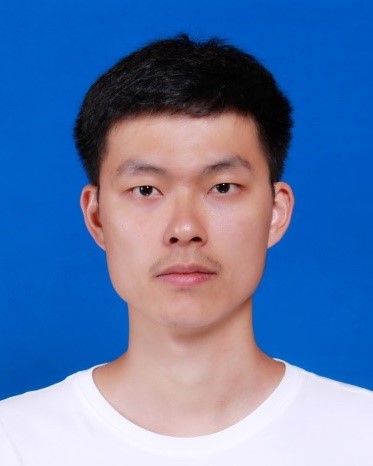}}]
	{Tianze Gao} received the B.S. degree in automation from the Harbin Institute of Technology, Harbin, China. Currently, he is working towards the Ph.D. degree in control science and technology at the Harbin Institute of Technology, Harbin, China. His research interests include autonomous driving technology, computer vision and artificial intelligence theories, with a particular focus on object detection and tracking in autonomous driving scenarios.
\end{IEEEbiography}
\vspace{-10 mm} 
\begin{IEEEbiography}[{\includegraphics[width=1in,height=1.25in,clip,keepaspectratio]{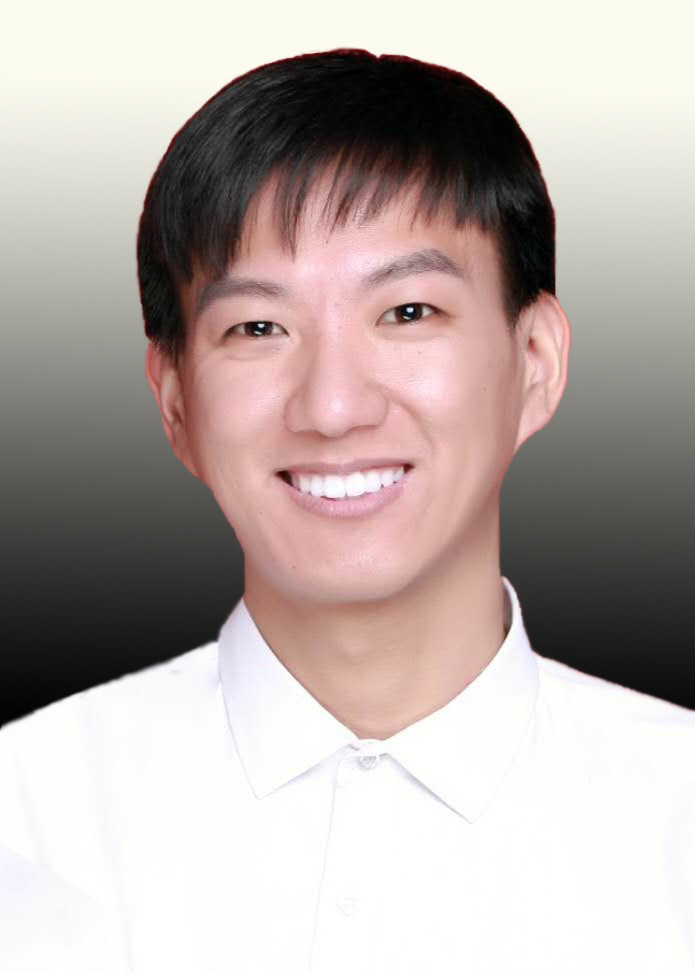}}]
	{Huihui Pan} received the Ph.D. degree in control science and engineering from the Harbin Institute of Technology, Harbin, China, in 2017, and the Ph.D. degree in mechanical engineering from The Hong Kong Polytechnic University, Hong Kong, in 2018. 
	
	Since December 2017, he has been with the Research Institute of Intelligent Control and Systems, Harbin Institute of Technology. His research interests include nonlinear control, vehicle dynamic control, and intelligent vehicles.
\end{IEEEbiography}
\vspace{-10 mm} 
\begin{IEEEbiography}[{\includegraphics[width=1in,height=1.25in,clip,keepaspectratio]{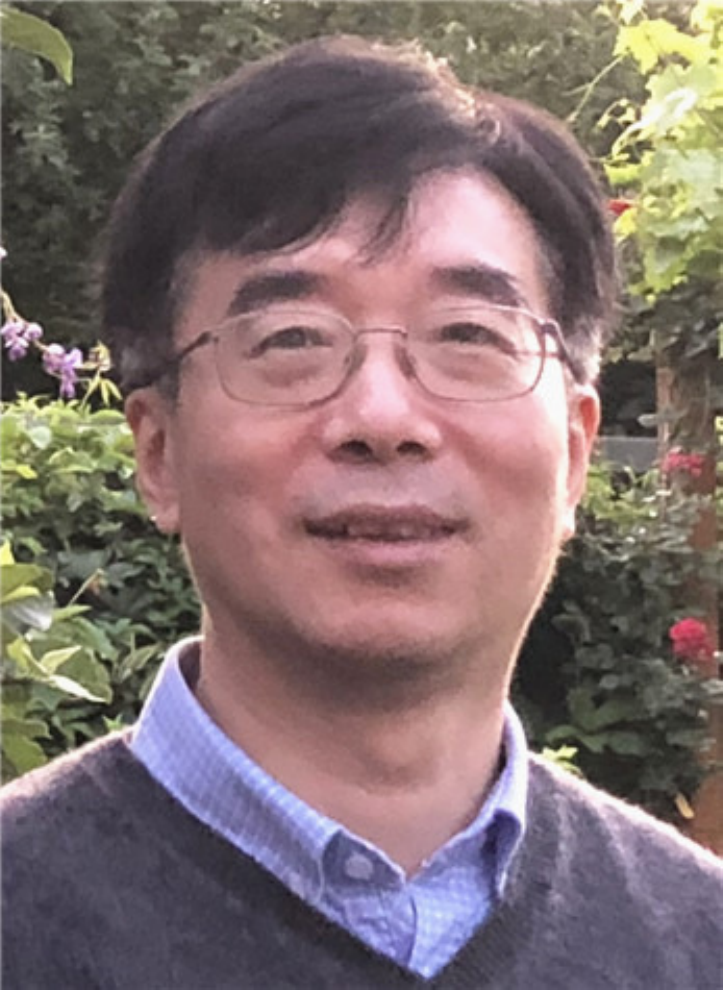}}]
	{Zidong Wang} (SM'03-F'14) was born in Jiangsu, China, in 1966. He received the B.Sc.~degree in mathematics in 1986 from Suzhou University, Suzhou, China, and the M.Sc.~degree in applied mathematics in 1990 and the Ph.D.~degree in electrical engineering in 1994, both from Nanjing University of Science and Technology, Nanjing, China.
	
	He is currently Professor of Dynamical Systems and Computing in the Department of Computer Science, Brunel University London, U.K. From 1990 to 2002, he held teaching and research appointments in universities in China, Germany and the UK. Prof. Wang's research interests include dynamical systems, signal processing, bioinformatics, control theory and applications. He has published more than 600 papers in international journals. He is a holder of the Alexander von Humboldt Research Fellowship of Germany, the JSPS Research Fellowship of Japan, William Mong Visiting Research Fellowship of Hong Kong.
	
	Prof.~Wang serves (or has served) as the Editor-in-Chief for {\it International Journal of Systems Science}, the Editor-in-Chief for {\it Neurocomputing}, and an Associate Editor for 12 international journals including \textsc{IEEE Transactions on Automatic Control}, \textsc{IEEE Transactions on Control Systems Technology}, \textsc{IEEE Transactions on Neural Networks}, \textsc{IEEE Transactions on Signal Processing}, and \textsc{IEEE Transactions on Systems, Man, and Cybernetics-Part C}. He is a Member of the Academia Europaea, a Fellow of the IEEE, a Fellow of the Royal Statistical Society and a member of program committee for many international conferences.
\end{IEEEbiography}
\vspace{-10 mm} 
\begin{IEEEbiography}[{\includegraphics[width=1in,height=1.25in,clip,keepaspectratio]{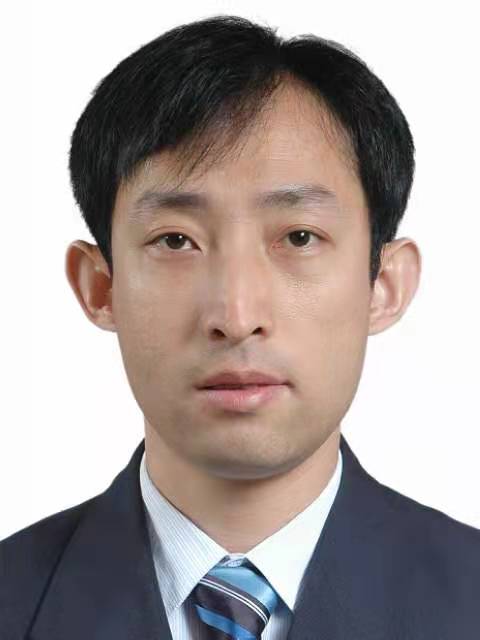}}]
	{Huijun Gao} (F'13) received the Ph.D. degree in control science and engineering from the Harbin Institute of Technology, Harbin, China, in 2005.
	
	From 2005 to 2007, he carried out his post-doctoral research with the Department of Electrical and Computer Engineering, University of Alberta, Edmonton, AB, Canada. Since 2004, he has been with the Harbin Institute of Technology, where he is currently a Full Professor, the Director of the Research Institute of Intelligent Control and Systems, and the Director of the Interdisciplinary Research Center. His research interests include intelligent and robust control, robotics, mechatronics, and their engineering applications. 
	
	Dr. Gao is an IEEE Industrial Electronics Society Vice President and a Council Member of the International Federation of Automatic Control (IFAC). He also serves as the Co-Editor-in-Chief for the \textsc{IEEE Transactions on Industrial Electronics}, a Senior Editor for the \textsc{IEEE/ASME Transactions on Mechatronics}, and an Associate Editor for \textit{Automatica}, the \textsc{IEEE Transactions on Cybernetics}, and the \textsc{IEEE Transactions on Industrial Informatics}.
\end{IEEEbiography}

\end{document}